\documentclass[runningheads,a4paper]{llncs}

\usepackage{amssymb}
\usepackage{graphicx}
\usepackage{caption}    
\usepackage{subfig} 
\graphicspath{ {./graphicspath/} }

\usepackage{url}
\urldef{\mailsa}\path|diegos.ulloa13@gmail.com|
\newcommand{\keywords}[1]{\par\addvspace\baselineskip
\noindent\keywordname\enspace\ignorespaces#1}

\begin{document}

\mainmatter  

\title{A Process for Topic Modelling Via Word Embeddings}

\titlerunning{A Process for Topic Modelling Via Word Embeddings}

%
%
\author{Diego Salda\~na-Ulloa}
%
\authorrunning{ }

\institute{
Puebla, M\'exico\\
\mailsa\\}

%
%

\toctitle{ }
\tocauthor{Authors' Instructions}
\maketitle

\begin{abstract}
This work combines algorithms based on word embeddings, dimensionality reduction, and clustering. The objective is to obtain topics from a set of unclassified texts. The algorithm to obtain the word embeddings is the BERT model, a neural network architecture widely used in NLP tasks. Due to the high dimensionality, a dimensionality reduction technique called UMAP is used. This method manages to reduce the dimensions while preserving part of the local and global information of the original data. K-Means is used as the clustering algorithm to obtain the topics. Then, the topics are evaluated using the TF-IDF statistics, Topic Diversity, and Topic Coherence to get the meaning of the words on the clusters. The results of the process show good values, so the topic modeling of this process is a viable option for classifying or clustering texts without labels.
\end{abstract}

\section{Introduction}

Text processing in the digital age is widely linked to daily processes occurring in all ambits. These types of tools allow obtaining several features related to the semantics of a text that can later be used for a wide variety of purposes, mainly related to the correct categorization of a document. The processes involving textual categorization are one of the tools with the most applications in different areas \cite{dieng}, considering the growing increase in information generation. This previous statement drives the need to automatically categorize a text (or document) as a fundamental task.

The process of categorizing a document or extracting its related topics is called topic modeling. Topic modeling involves collecting a set of latent variables that help define concepts from documents \cite{vayansky}. This type of process can be helpful in sites with a large amount of information, for example, in databases of journals or articles, and even in content generated in social networks by users \cite{blei,hong}.

The methods to extract topics from a document look for a way to access the semantic information of the text; in this way, different types of processes that combine techniques and statistical algorithms can be designed. As a first step, using processes based on the count of words is common. However, these techniques could not consider the dependency between the document's different terms \cite{ramage}. This is the reason for the necessity of techniques that consider both the words' context and the frequency of terms.

In this work, we propose a set of techniques and algorithms that consider the latent variables of a document, the context of the words (through word embeddings), and the frequency of the different terms present in a text to perform topic modeling. The organization that will be followed will be the one described below: Section 1 describes the background related to the topic modeling area as well as the related works, section 2 introduces the theory of the different methods and algorithms used, section 3 presents the description of the process for the extraction of topics from a document, in section 4 the experimental results are detailed, and finally, the conclusions are handled.

\section{Background and Related Works}

Topic modeling is a set of techniques used to extract information from a document and define a set of ideas that characterize it. There are different methods to carry out this process, but many arise from the Vector Space Model (VSM) \cite{kherwa1}. VSM is a mathematical model to represent texts that work considering the relevance (numerical weights) of different terms on a set of documents \cite{bookt}. Relevance is generally assigned by a function related to the frequency of each term in the entire document. In this way, an n-dimensional vector type can be defined for each document, formed by the numerical weights of each word.

TF-IDF is one of the first methods used to consider the relevance of a term over the whole document. This method considers the frequencies of the terms (TF) over the total corpus size (IDF). The result of this procedure is precisely a matrix of weights for each of the terms in a document. The TF-IDF method is used by another technique called Latent Semantic Analysis (LSA) which is used in natural language processing tasks such as text classification. LSA factors the TF-IDF matrix using Singular Value Decomposition (SVD) and thus manages to reduce its dimensionality. In this process, part of the semantic information of the terms is captured, so LSA can be used to assign topics to a document based on the numerical weights of each term over the entire text \cite{alghamdi}.

Another method that is used in topic extraction is PLSA (Probabilistic Latent Semantic Analysis). It is based on LSA, and they differ in that PLSA considers that each term comes from a probability distribution given for each one of the topics. In this way, a set of probability distributions of a fixed set of topics characterizes each document. The model estimates the topic and word distributions that best explain a co-occurrence of terms in the \cite{vayansky} corpus.

Latent Dirichlet Allocation (LDA) is another method used for topic modeling. It works similarly to PLSA, each document is considered a mixture of topics, and each topic comes from a probability distribution of the possible words in the topic. The algorithm works by assigning words to topics and iteratively refining the assignment by maximizing the probability of the data. This is done through a Bayesian estimate of the probability distributions of words and topics given the observed data \cite{alghamdi}.

The methods described above are commonly used for topic modeling tasks. Several works use them as part of a text categorization process or through comparing methods \cite{alghamdi,Kalepalli,george,Nallapati}. However, LSA, PLSA, and LDA have several disadvantages, such as fixing the number of topics, lack of capturing non-linear relationships between words, and the assumption that a document can contain different topics, which may not be valid. For this reason, other proposals have been developed that consider the possible dependence on the context of words.

Deep learning is today's most widely used method for natural language processing tasks. Different models and architectures can capture the dependency between the words in a sentence. With this consideration, the task focuses on obtaining an n-dimensional representation of a word or sentence, i.e., an embedding. The use of word embeddings has become increasingly widespread nowadays. The advantage of an n-dimensional representation is that it can be used from different approaches \cite{dieng}.

Some works have focused on using word embeddings for topic modeling tasks \cite{bertopic,Suhyeon,moody}. They generally work by obtaining the embeddings for the texts of each document and then applying term frequency techniques, clustering, and combinations with LDA. In the present work, this approach is used, combining the extraction of word embeddings through a pre-trained BERT (Bidirectional Encoder Representations from Transformers) model, a UMAP dimensionality reduction technique \cite{mcinnes}, and the clustering of the elements as the final step to get a set of topics using K-means. Some methods and results of \cite{bertopic} inspired this work.

\section{Word Embeddings, Dimensionality Reduction, and Topic Modelling} 
Word embedding represents a word or text as an n-dimensional vector considering the semantic and syntactic characteristics within the corpus \cite{yang}. This type of representation is commonly used in multiple NLP tasks, generally as features that feed some machine learning algorithm. The most usual way to obtain word embeddings is through different types of neural network architectures. It has been verified that better results are obtained using neural networks than techniques based on the frequency of terms \cite{almeida}.

\subsection{Transformers and BERT Model for Word Embedding}

One of the most used architectures currently in NLP tasks is the Transformers, particularly a derived model called BERT. Transformers are a type of deep neural network architecture that uses attention mechanisms (weights between input data elements) to select the most important parts of an input sequence, catching the long-range dependencies of a sentence \cite{Gillioz}. They work through an encoder-decoder structure that takes $X=\{x_1, x_2, ..., x_N\}$ sequences (tokenized values from the input text) and produces $X=\{z_1, z_2, ..., z_N\}$ representations. The encoder-decoder blocks use multi-head attention mechanisms to capture the dependencies between different tokens in the input sequence \cite{Vaswani}. In general, this type of architecture uses an stack of encoder-decoder layer stacks; that is, the information is autoregressive and depends on the previous results of the computation.

BERT comprises a set of Encoder Transformer layers and an attention mechanism to capture relationships between all the words in a sentence. The difference from BERT is that this unit deals with the forwards and backward of a sentence \cite{Devlin}. This model is pre-trained on a large text corpus using two unsupervised learning tasks: masked language modeling and next-sentence prediction \cite{Gillioz}. The first part consists of predicting the original masked tokens (tokens generated on the input data) based on the context of the surrounding words. This allows the model to understand relationships between words in a sentence. In the next sentence prediction, the model aims to predict if a pair of sentences follow each other. This helps BERT to understand the relationship between sentences in a document. Since pre-training aims to learn dependencies and relationships between words and sentences, the model can be fine-tuned to perform multiple tasks as word embedding \cite{Vaswani}.

\subsection{UMAP for Dimensionality Reduction}

The dimensionality reduction plays an essential role in data analysis and visualization tasks. Some of these techniques work through multiple transformations on the input data that are linear combinations of the initial information. Dimensionality reduction techniques generally do not consider the global characteristics of the input data, so the distance of a set of points in the original space will not necessarily be preserved in the reduced space \cite{Vermeulen}.

Uniform Manifold Approximation and Projection (UMAP) is a dimension reduction technique that uses graph layout algorithms to arrange data in a low-dimensional space \cite{mcinnes}. The main idea focuses on approximating the high-dimensional data manifold (curved surface embedded in a high-dimensional space). The embedding of the data is computed by searching for a low-dimensional projection of the data with the closest possible equivalent global shape and structure as the original dataset \cite{Vermeulen}, i.e., UMAP preserves the local and global structure of the data. This method considers that there is a manifold on which the data would be uniformly distributed, and the main objective is to preserve the topological structure of this manifold \cite{mcinnes}.

\subsection{K-means Clustering}

The $K$-means algorithm is one of the best-known methods for solving clustering problems. $K$-means operates on a set of observations to try to group them into a certain number of sets, considering that the square of the distance between each element and its cluster center is the minimum possible \cite{arthur}. We can formally define the $K$-means problem as the next optimization problem.

$K$-means optimization problem: Given a set of elements $\textbf{x}_n \in X$, $X \subset R^D$ 
and an integer $k \leq n$ such that there are $c_k \in C$ subsets. The objective of the $K$-means clustering is to partition $n$ elements into $k$ sets $S$ to minimize the within-cluster sum of squares (WCSS) i.e.

\begin{equation}
    \min_{c \in C} \sum_{x \in X} \left\Vert  x-c \right\Vert^2.
\end{equation}

During each iteration of the algorithm, the following steps are computed: first, the $k$ cluster centroids are taken randomly from the elements of $X$. Next, each data element is assigned to its nearest cluster. New centroids are created by taking the average value of the elements assigned to the cluster. Finally, the difference between the new centroid and the previous one is calculated. The algorithm stops until there are no significant differences between these two values.

\begin{figure}[h]
\includegraphics[scale=0.48]{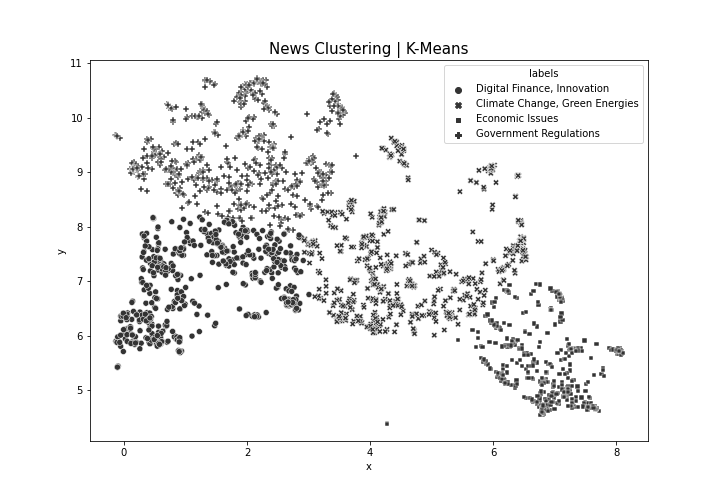}
\caption{Clustering of the reduced embeddings}
\label{fig:emtwo}
\end{figure}

\section{Topic Modelling with Word Embeddings}

The process used for topic modeling consists of multiple of steps involving the algorithms described above. A pre-trained BERT model \cite{Reimers2019SentenceBERTSE,reimers-gurevych-2020-making} is used as a first step to obtain the embeddings of a set of texts. Using a pre-trained model has an advantage, such as time and resource efficiency or a broad understanding of the language, because these models are trained on a large amount of text data. Since the pre-trained BERT models to generate embeddings result in high-dimensionality vectors \cite{Devlin}, it is necessary to use dimensionality reduction techniques. That is why as a second step, the UMAP algorithm is applied. The reduced dimensions preserve the original data's local and global structure through nonlinear transformations \cite{mcinnes}. BERT generates embeddings considering the context of the words in the corpus; therefore, similar text strings must have similar embeddings in the final space. However, difficulties have been observed in vectors with high dimensionality when using distance metrics \cite{Devlin}.

For this reason, using dimensionality reduction techniques such as the one described above is expected. By applying UMAP, this information manages to be preserved due to how the algorithm operates. To this point, generating embeddings with a reduced dimension for the text set to be described is possible. An additional step is needed to use this information to generate topics from a set of documents. The third step of the process described in this work involves using clustering techniques to assign clusters to the embeddings obtained. K-means is used in this work due to its ease of use and interpretability. The clusters obtained by K-Means on each document's set of embeddings correspond to each text's topics.

\begin{figure}[h]
\includegraphics[scale=0.7]{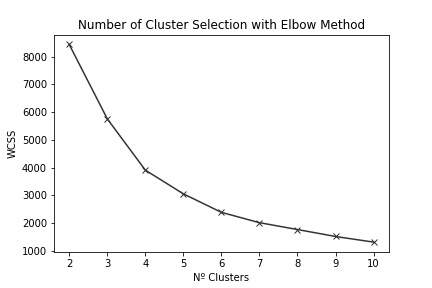}
\caption{Elbow method to select the number of clusters}
\label{fig:emethod}
\end{figure}

\section{Experimental Results}

A dataset of 1212 news items in Spanish extracted from a website \cite{kaggle} was used for the experimental process. As part of the preprocessing, punctuation marks and special characters ($\$, \#, \$, \%, \&$, among others) were removed. Likewise, common words without significant relevance (stopwords) from Spanish were eliminated in concordance with \cite{booknltk}. For this process, a pre-trained BERT model was used in more than 50 languages, including Spanish. The BERT model, pre-trained in more than 50 languages, is designed to preserve the distance between words with the same meaning in different languages \cite{Reimers2019SentenceBERTSE,reimers-gurevych-2020-making}. This model is used because of the limited availability of BERT models trained exclusively in Spanish.
Because of the characteristics of the pre-trained model (with English and Spanish languages included), a random sample composed of half of the news was taken, then its translation into English was obtained. This translated sample was incorporated into the original Spanish dataset to assess whether the resulting topics would contain similar words from both languages. The final size of the dataset was 2183 text news.

Once this process was completed, the embeddings were obtained (with the pre-trained BERT model in more than 50 languages) for each news text. The resulting embeddings had a dimensionality of 768, according to what was reported in the literature \cite{Devlin}. Subsequently, the UMAP dimensionality reduction algorithm was used, preserving two dimensions. The reasons for this choice focused on having a direct relationship between the two-dimensional visualization and the subsequent results, figure \ref{fig:emtwo}. The K-Means algorithm was used as a third step to obtain clusters on the embeddings of the texts. The heuristic called the elbow method was used to select the number of clusters \cite{thorndike}. The initial centroids required by the method were chosen according to the initialization of the K-Means++ algorithm \cite{kmeansplus}; this guaranteed that the distance between the initial centroids was distant, which ensured a higher probability of better results. In addition, the method was repeated 100 times, and the best result was selected according to \cite{kmeansbestinit}. Figure \ref{fig:emethod} shows the result of the elbow method applied; it is observed that the best number of clusters is four.

With the clusters obtained, the words of each text were grouped, and the TF-IDF statistic was applied. TF-IDF is a statistic used to evaluate the importance of a term in a document considering a corpus of words \cite{bookt}. To calculate it, the term frequency TF is used, which is an estimate of the probability of occurrence of a term in the document \cite{aizawa}, and the inverse document frequency (IDF), which is the change in the amount of information of a term above all the corpus \cite{aizawa} or the rarity of the term considering the corpus. TF-IDF is the product of TF and IDF components.

\begin{figure}[h]
\includegraphics[scale=0.44]{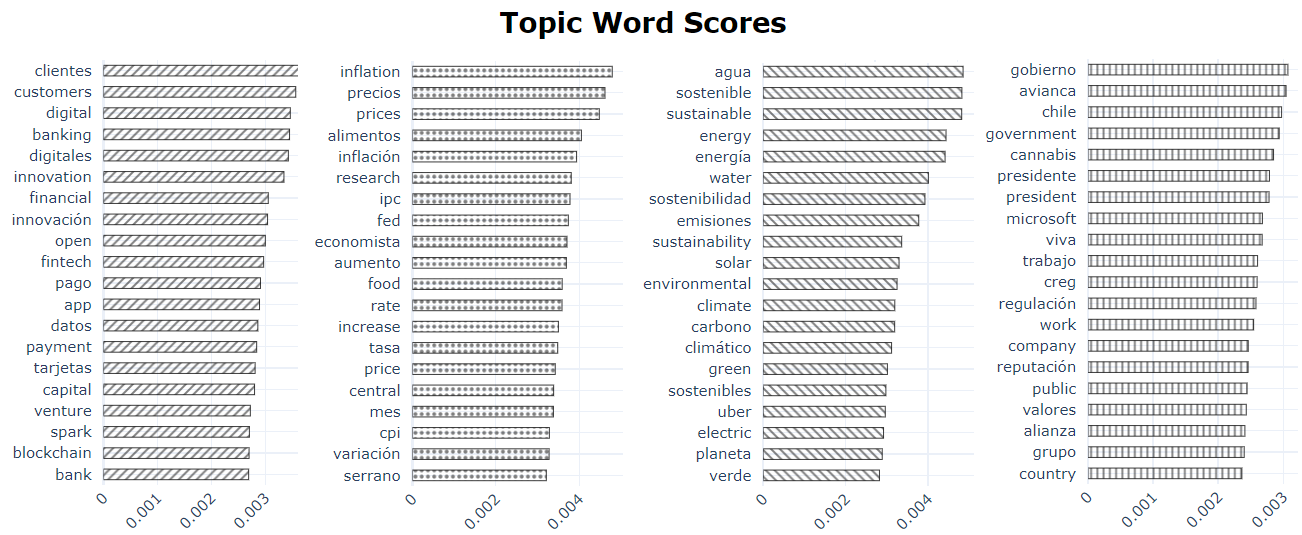}
\caption{Topics obtained with K-Means. The words and their TF-IDF score are shown.}
\label{fig:topicskmeans}
\end{figure}

With this process, the most important words for each cluster can be obtained. This behavior is shown in \ref{fig:topicskmeans}. The clustering carried out by K-Means does an excellent job since the words seem related in their meaning. The results should also be attributed to the embeddings obtained by BERT and to the dimension reduction process with UMAP that manages to preserve the global and local information of the original BERT vector. Additionally, it is observed that some topics contain the same words but in a different language due to the mentioned characteristics of the BERT, model pre-trained in more than 50 languages. Based on these results, topic 0 corresponds to subjects related to digital finance and innovation, topic 1 to economic issues, topic 2 to climate change and alternative energy, and topic 3 to types of government regulations or government agreements.

To statistically evaluate the results obtained by the combination of methods, two metrics extracted from \cite{bertopic} Topic Diversity and Topic Coherence were used. Topic Diversity is used to evaluate the redundancy of words in each document, that is, the number of unique terms with respect to a document or text. The range of values is between 0 and 1, where 1 indicates a great diversity of terms and 0 is more redundant terms or words \cite{bertopic}. Topic Coherence is a metric used to measure the association of terms in a corpus. It works by calculating the probabilities of occurrence of two terms in the same document \cite{Bouma2009NormalizedM}; this is done for all the terms of the same topic considering the corpus. The range of values goes from [-1,1], where 1 indicates a perfect association between the terms; that is, there is a co-occurrence (coherence) of the terms in the document.

\begin{figure}[ht]
\includegraphics[scale=0.43]{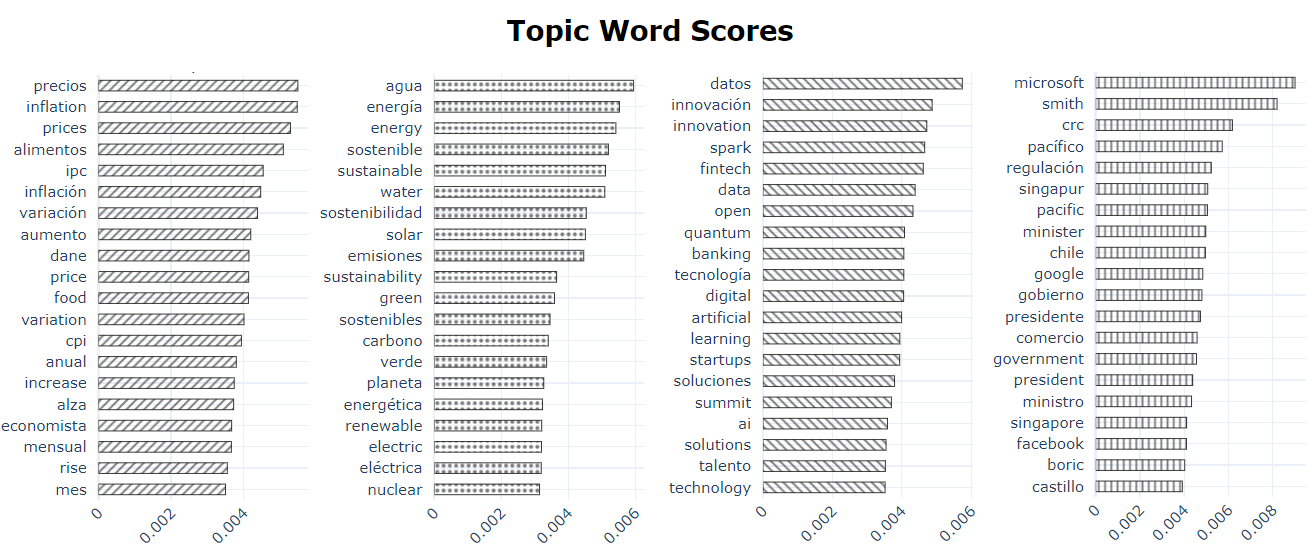}
\caption{Topics obtained with HDBSCAN. The words and their TF-IDF score are shown.}
\label{fig:topicshscan}
\end{figure}

Table \ref{table:comparison} show the results of Topic Diversity and Topic Coherence for the process described in this work. A comparison with the method described in \cite{bertopic} is also shown. A significant difference with \cite{bertopic} is that the author of that work used a modified version of the DBSCAN clustering technique called HDBSCAN as the algorithm to get the clusters of the embeddings of the texts.  Both methods show similar results regarding the Topic Diversity and Topic Coherence metrics. It is also observed that the general context of each topic is similar for both clustering techniques. The parameters selected for HDBSCAN were such that four topics were obtained (just like K-Means). In addition, figure \ref{fig:topicshscan} shows the topics created by the HDBSCAN algorithm.

\begin{table}[ht!]
\caption{Comparison between clustering methods}
  \begin{tabular}{cccc}
    \hline
  Cluster method & Topics & Topic Diversity & Topic Coherence  \\ \hline
  \\
     & Digital finance, Innovation & 0.25 & 0.53  \\
   K-Means  & Economic issues & 0.3 & 0.50  \\
     & Climate change, green energies & 0.35 & 0.31  \\
     & Government regulation & 0.425 & 0.11  \\
     \hline 
     \\
     & Digital finance, technology & 0.24 & 0.46  \\
   HDBSCAN  & Economic issues& 0.32 &  0.4  \\
     & Climate change, green energies & 0.38 & 0.31  \\
     &  Government regulation & 0.4 &  0.16  \\
   
   \\
      \hline 
  \end{tabular}
\label{table:comparison}
\end{table}

\section{Conclusions}

This work combined several algorithms to obtain the topics for a Spanish news dataset. The text was preprocessed, eliminating Spanish stopwords, punctuation marks, and special characters. Likewise, a random sample of half of the dataset was taken and translated into English, and this sample was incorporated into the original dataset. The later was done due to the type of model used to get the text embeddings. With the pre-processed text, a pre-trained BERT model trained on more than 50 languages was used to obtain each text's embeddings. Due to the high dimensionality of the resulting vectors, the UMAP dimensionality reduction technique was applied.

Subsequently, the K-Means algorithm was used to obtain a defined number of clusters. In this case, the number was selected by applying the cubit method. To visualize which sets of words were found within each cluster, the TF-IDF statistic was applied to obtain the most relevant words for each topic. Finally, two metrics related to the diversity of terms and coherence were applied to the topics found. These results were compared with those obtained in a previous work \cite{bertopic}. A good relationship is observed between the topics and the coherence and diversity of the terms.

With the process described in this work, good results are obtained in the topic modeling of a set of texts. The described technique could be used in another type of process in real applications where it is necessary to classify a set of unthematic texts.

\end{document}